\documentclass{article}

\usepackage{amsmath}
\usepackage{amsfonts}
\usepackage{amsthm}
\usepackage{microtype}
\usepackage{marginnote}
\usepackage{xcolor}
\usepackage{enumitem}   
\usepackage{multirow,tabularx}
\usepackage{subfigure}
\usepackage{tikz}
\usepackage{dialogue}
\usepackage{appendix}
\usepackage{breakcites}
\usepackage{booktabs}
\usepackage{listings}
\usepackage{graphics}
\usepackage{graphicx}
\usepackage[framemethod=TikZ]{mdframed}
\usepackage[hidelinks]{hyperref}
\usepackage[letterpaper,margin=1.75in]{geometry}
\usepackage[font={small,it}]{caption}

\graphicspath{{./figures/}}

\let\citep\cite
\let\citet\cite

\usepackage{expl3}[2012-07-08]
\ExplSyntaxOn
\ExplSyntaxOff

\DeclareMathOperator*{\argmin}{arg\,min}

\usetikzlibrary{decorations.pathmorphing,shapes}

\title{Counterbalancing Teacher: Regularizing Batch Normalized Models for Robustness}
\author{Saeid Asgari Taghanaki$^{*\text{1}}$, Ali Gholami$^{*\text{1}}$, Fereshte Khani$^{*\text{2}}$,\\ Kristy Choi$^\text{2}$, Linh Tran$^\text{1}$, Ran Zhang$^\text{1}$, Aliasghar Khani$^\text{1}$}
\date{$^1$Autodesk AI Lab, $^2$Stanford University}

\begin{document}

\maketitle
\def\thefootnote{*}\footnotetext{These authors contributed equally to this work}\def\thefootnote{\arabic{footnote}}

\begin{abstract}

Batch normalization (BN) is a ubiquitous technique for training deep neural networks that accelerates their convergence to reach higher accuracy.
However, we demonstrate that BN comes with a fundamental drawback: it incentivizes the model to rely on low-variance features that are highly specific to the training (in-domain) data,
hurting generalization performance on out-of-domain examples. In this work, we investigate this phenomenon by first showing that removing BN layers across a wide range of architectures leads to lower out-of-domain and corruption errors at the cost of higher in-domain errors. We then propose Counterbalancing Teacher (CT), a method which leverages a frozen copy of the same model without BN as a \textit{teacher} to enforce the student network's learning of robust representations by substantially adapting its weights through a consistency loss function. This regularization signal helps CT perform well in unforeseen data shifts, even without information from the target domain as in prior works. We theoretically show in an overparameterized linear regression setting why normalization leads to a model's reliance on such in-domain features, and empirically demonstrate the efficacy of CT by outperforming several baselines on robustness benchmarks such as CIFAR-10-C, CIFAR-100-C, and VLCS.

\end{abstract}
\section{Introduction}
\label{intro}

Batch normalization (BN), a neural network layer that 
% mitigates internal covariate shift 
normalizes input features
by aggregating batch statistics during training, is a key component for accelerating convergence in the modern deep learning toolbox~\citep{ioffe2015batch,santurkar2018does,bjorck2018understanding}. It plays a critical role in stabilizing training dynamics for large models optimized with stochastic gradient descent, and has since spurred a flurry of research in related extensions~\citep{ba2016layer,kingma2014adam} and its understanding~\cite{gitman2017comparison,santurkar2018does,luo2018understanding, kohler2018towards}.

Despite its advantages, BN has recently been shown to be a source of vulnerability to adversarial perturbations~\citep{galloway2019batch,benz2021revisiting}.
In our work, we take this observation one step further and demonstrate that BN also compromises a model's out-of-domain (OOD) generalization capabilities. 

Specifically, we demonstrate that normalization incentivizes the model to exploit highly predictive, low-variance features~\citep{geirhos2018generalisation,geirhos2020shortcut}, that lead to poor classification accuracy when the test environment differs from that of training.
Given the widespread use and benefits of normalization, we desire a way to mitigate such drawbacks in models trained with BN.

To better understand this phenomenon, we investigate the effect of normalization in over-parametrized regimes, where there exist multiple solutions and inductive bias (e.g., minimizing the norm of the weights) significantly impacts the estimated parameters. 
Similar to recent work in the theory of deep learning~\citep{khani2021removing,raghunathan2020understanding,nakkiran2019more, hastie2019surprises,liang2020just}, we study the minimum-norm solution in over-parametrized linear regression. 
Without normalization, the inductive bias selects a model that fits training data and minimizes a fixed norm independent of data; with normalization,
the same inductive bias selects a model that minimizes a data-dependent norm, leading the model to rely more on low-variance features.
While such highly predictive features yield better performance in-domain where the features do not vary significantly, they cause performance to plummet in OOD settings (e.g. data corruptions or missing features).
This is in direct contrast to models trained without BN that assign equal weight to \emph{all} input features, which help to reduce their overfitting on the training set.

Drawing inspiration from our observation and the knowledge distillation literature~\citep{hinton2015distilling,romero2014fitnets}, we propose a simple student-teacher model to combine the best of both worlds.
That is, we leverage features derived \emph{both} from a network without BN (teacher) and its clone with BN (student) to train a model to learn representations that achieve high accuracies in standard and robust settings.

\begin{figure}[t!]
     \centering
     \includegraphics[width=0.75\textwidth]{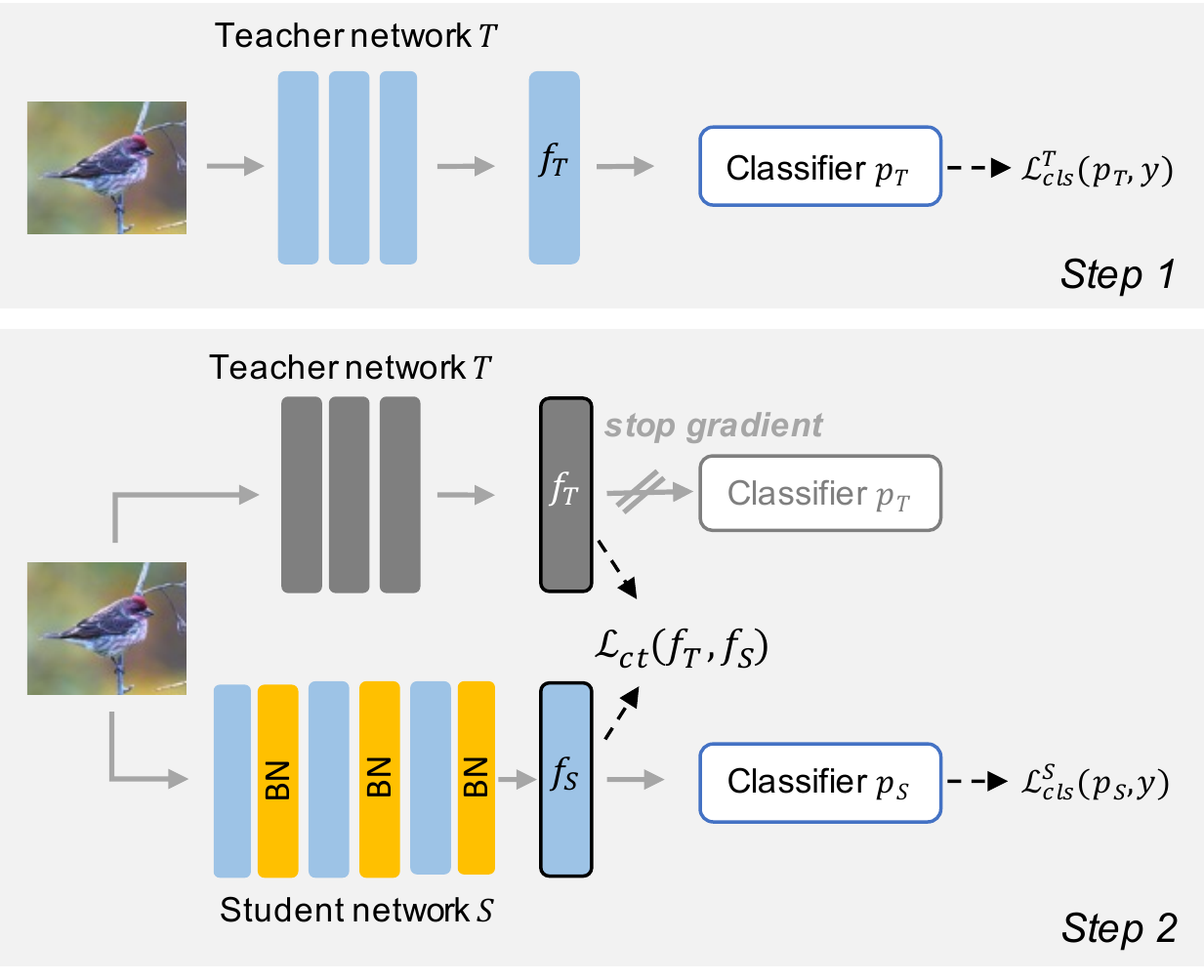}
     \caption{\textbf{The Counterbalancing Teacher (CT) model overview.} The optimization consists of two steps. First, we train the the teacher model (without any BN layers) by minimizing a cross-entropy objective loss. Once the teacher is fully trained, we freeze its weights. Second, we train the student model (with BN layers) using the cross-entropy loss and distill the teacher's knowledge by regularizing the student's and teacher's representations $f_S$ and $f_T$. During inference, we only use the student model. The grey and blue colours represent frozen and unfrozen models, respectively.}
     \label{fig:method}
 \end{figure}

We incorporate a regularization term in the loss function which encourages the features learned from the student encoder to have similar statistics and structure to those learned from the teacher; we name this model the Counterbalancing Teacher (CT) and show that it helps in achieving both higher robust and clean accuracy compared to a (batch) normalized model. 
In particular, CT retains good performance in OOD settings even \emph{without} knowledge of statistics of the new domain.
We provide an illustrative flowchart of the CT framework in Figure~\ref{fig:method}.

\looseness=-1
Our results mark a significant improvement over prior works, which have 
tackled similar problems by either: (a) modifying batch normalization statistics of a trained model~\citep{schneider2020improving,benz2021revisiting} using privileged information from the target domain; or (b) augmenting the training data using a set of predefined corruption functions~\citep{hendrycks2019augmix}. 
As recent studies~\citep{vasiljevic2016examining,geirhos2018generalisation,taghanaki2020robustpointset} show that such approaches often fail to generalize due to the tendency of neural networks to memorize data-specific properties, this motivates a shift towards developing models that are inherently robust, regardless of particular data augmentations or input transformations applied during training.

\looseness=-1
Empirically, we demonstrate that CT outperforms most existing data augmentation-based techniques and covariate shift adaption-based methods~\citep{hendrycks2019benchmarking} (which require information from the test set) on mean corruption error on CIFAR10-C and CIFAR100-C~\citep{hendrycks2019benchmarking}.
CT also achieves state-of-the-art performance in domain generalization on the VLCS dataset~\citep{torralba2011unbiased}. We further test CT on corrupted 3D point-cloud data~\citep{taghanaki2020robustpointset} and show it outperforms existing methods in terms of mean classification accuracy over multiple test sets. To the best of our knowledge, this is the first work to explore both theoretically and empirically why BN leads to a model's over-reliance on brittle, low-variance features, which can adversely affect its performance on the downstream classification task.
This is also the first work to present a robust representation learning framework for common input distortions without additional data augmentation strategies or information derived from the target domain.
In summary, our contribution is threefold:
\begin{enumerate}
    \item We justify theoretically why normalization encourages a model to exploit low-variance features, and empirically evaluate how this behavior can adversely affect downstream classification accuracy. 
    \item We propose CT, a representation learning approach demonstrating that regularizing representations of a batch-normalized network (student) using those from an unnormalized copy of the network (teacher) can significantly improve the model's robustness. 
    \item We experimentally verify the robustness of the representations learned by CT to input distortions and domain shifts on a variety of tasks and models. 
\end{enumerate}

\section{Related Work}
\label{related}

\textbf{Reducing classification error on common input corruptions.} ~\cite{hendrycks2019augmix} proposed a data augmentation technique which mixes multiple augmented samples using random weights and showed improvements on robust accuracy.~\cite{rusak2020simple} proposed to mix Gaussian noise with adversarial samples for robustness to common distortions. However, as in the real world, the space of distortions and their mixture is not finite, it is not trivial to devise a set of augmentation types that will make a model robust to all distortions. Moreover, making a model robust to certain corruptions via augmentation does not generalize to others.~\cite{dodge2017quality}
leverage an ensemble approach for robustness against distortions, but they also assume corruptions are known beforehand.\\
\\
\textbf{Batch normalization's vulnerability.} Several works have studied the effect of batch normalization in the context of robustness to adversarial perturbations. However, as the most relevant works to ours, ~\cite{schneider2020improving} and~\cite{benz2021revisiting} suggested improving robustness to common data corruptions by updating the batch normalization parameters using statistics calculated form extra test distribution samples. However, extra test samples are not always available, and similar to data augmentation-based methods, these approaches work only when a model is adapted to a specific distortion and tested on the same distortion type.\\
\\
\textbf{Teacher-student models for robustness.} The focus of the few existing methods on leveraging teacher-student models for robustness has been on adversarial perturbations for developing adversarially robust models either by training teacher encoders with adversarial~\citep{goldblum2020adversarially,papernot2016distillation} or augmented~\citep{arani2021noise} examples. In contrast, our CT method is augmentation-independent. Adversarially trained models might not necessarily be robust to common data corruptions, as we showed in Table~\ref{tab:data_augs}, our CT method performs significantly better (at least by 12\%) on both CIFAR-10-C and CIFAR-100-C datasets across different models compared to the adversarially trained model (AdvT). Even if the augmented samples are crafted by common data augmentations rather than adding adversarial noise, the robustness obtained by the augmentations does not generalize to unseen corruptions, even when they are from the same family of the augmentations a model has seen~\citep{vasiljevic2016examining} during training.

\section{Problem Statement and Analysis}
\label{theory}
\looseness=-1
We assume a supervised classification setting: given a set of inputs $\mathcal{X} \subseteq \mathbb{R}^d$ and their corresponding labels $\mathcal{Y} = \{1, ..., k\}$,
we aim to learn a classifier $f_{\zeta}: \mathcal{X} \longrightarrow \mathcal{Y}$ by minimizing the empirical risk: 
\begin{equation}
\begin{split}
  \zeta &= \operatorname*{argmin}_\zeta \mathbb{E}_{x, y \sim p_{d}(x, y)}[\ell(x, y; \zeta)] \approx \operatorname*{argmin}_\zeta \sum_{i=1}^{n} \ell(f_{\zeta}(x_i), y_i) 
\end{split}
\label{eqn:nn_real_model}
\end{equation} 
where $p_d(x, y)$ denotes the underlying joint distribution over the labeled dataset.

In the following sections, we first discuss the effect of (batch) normalization on the solutions found in underspecified regimes, then elaborate on our approach to optimize the aforementioned empirical risk. In this context, we refer to a problem as underspecified or overparameterized when degrees of freedom of a model $f$ is larger than the number of training examples.

\subsection{The effect of normalization in overparametrized regimes}
\label{overparam}
\looseness=-1
Modern deep learning frameworks usually incorporate many parameters (often larger than the number of training data points).
% , which lead to underspecified regimes.
In other words, many distinct solutions solve the problem equally well i.e., have the same training or even held-out loss ~\citep{d2020underspecification}.
In this underspecified regime, the inductive bias of the estimation procedure, such as choosing parameters with the minimum norm, significantly impacts the estimated parameters. 
In such regimes, we show that normalizing data incentivizes the model to rely on features with lower variance. 
We analyze the effect of normalization on the min-norm solution in overparametrized noiseless linear regression. This setup has been studied in many recent works for understanding some phenomena in deep networks~\citep{khani2021removing,raghunathan2020understanding,nakkiran2019more}.

Let $X \in \mathbb{R}^{n\times d}$ denote training examples and $Y \in \mathbb{R}^{n}$ denote their target. 
As we are in an overparametrized regime ($d >n$), there should be an equivalence class of solutions. %are many $\theta \in \mathbb{R}^d$ such that $X\theta = y$. 
We assume that the inductive bias of the model is to choose the min-norm solution (the parameter with the minimum $\ell_2$ norm). 
This is in line with the recent speculation that the inductive bias in deep networks tends to find a solution with minimum norm~\citep{gunasekar2018implicit}. One can show that the convergence point of gradient descent run
on the least-squares loss is the min-norm solution.

Without normalization, the model chooses the min-norm solution which fits the training data:
\begin{align}
\label{eqn:nn}
   \hat \zeta =  \argmin_\zeta \|\zeta\|_2^2 \quad  s.t. \quad X\zeta = Y.
\end{align}
\looseness=-1
Now we observe how normalization changes the estimated parameters.
Let $U$ be a diagonal matrix where $U_{ii}$ denotes the standard deviation of the $i^\text{th}$ feature. 
By normalization, we transform $X$ to $XU^{-1}$ (for simplicity, we assume that the mean of each feature is $0$ and can show that transforming the points does not change the estimated parameter, see Appendix~\ref{sec:subtracting_by_mean_is_ok} for details).
In this case, the model estimates $\hat\beta$ as follows:
\begin{align}
\label{eqn:helper}
    \hat\beta = \argmin_\beta \|\beta\|_2^2 \quad
    s.t. \quad XU^{-1}\beta = Y,
\end{align}
and since we normalize data points at the test time as well, the estimated parameter used for prediction at the test time is  $\hat \theta = U^{-1}\hat\beta$.
Substituting $\theta$ instead of  $U^{-1}\beta$ (thus $U\theta = \beta$), we can write the equal formulation of \ref{eqn:helper} as: 
\begin{align}
\label{eqn:wn}
    \hat\theta = \argmin_{\theta} \|U\theta\|_2^2 \quad
    s.t. \quad X\theta = Y.
\end{align}
For the same equivalence class of solutions, a model with normalization (Eq.~\eqref{eqn:wn}) chooses different parameters in comparison to a model without normalization (Eq.~\eqref{eqn:nn}).
In particular, Eq.~\eqref{eqn:nn} chooses an interpolant with a minimum \emph{data independent} norm.
On the other hand, Eq.~\eqref{eqn:wn}, chooses an interpolant with a minimum \emph{data-dependent} norm, which incentives the model to assign higher weights to low variance features. 
Note that projection of $\hat \theta$ and $\hat \zeta$ is the same in column space of training points. 
Formally if $\Pi = X^\top (XX^\top)^{-1}X$ denotes the column space of training points then $\Pi \hat \theta = \Pi \hat \zeta$.
However, their projections to the null space of training points ($I-\Pi$) are different.
Consequently as we have more data (and thus a smaller null space), $\hat \theta$ and $\hat \zeta$ become closer, and converge when $n > d$. 
We note that our analysis holds for classification with max-margin, where we only need to substitute $X\theta = Y$ by $Y \odot X\theta \ge 1$. 

\subsection{The effect of normalization on max-margin classifiers}
\label{sec:max-margin}
In Section~\ref{overparam}, we analyzed the effect of normalization in overparameterized linear regression. Here we show that the analysis holds for the max-margin classifiers as well. 
\cite{soudry2018implicit} show that without any explicit regularization, gradient descent on the logistic loss converges to the L2 maximum margin separator for all linearly separable datasets. 
In the overparametrized regime where data are completely separable, there is also some investigation that as $d$ increases, all the data points serve as support vectors \cite{narang2020overparameterized}. Without normalization we have:
\begin{align}
    \hat \zeta = \argmin \|\zeta\|_2^2\quad \quad
    s.t. \quad  Y \odot X \zeta \ge 1
\end{align}
Recall that $U$ is a diagonal matrix where $U_{ii}$ is the standard deviation of the $i^\text{th}$ feature. After normalization we have:
\begin{align}
    \hat \beta = \argmin \|\beta\|_2^2\quad \quad 
    s.t. \quad Y \odot XU^{-1} \beta \ge 1
\end{align}
At the test time we predict $\text{sign}(XU^{-1} \hat \beta)$, substituting $U^{-1}\beta$ with $\theta$ we have:
\begin{align}
    \hat \theta = \argmin \|U\theta\|_2^2\quad \quad 
    s.t. \quad Y \odot X\theta \ge 1
\end{align}
Similar to the linear regression setup, normalizing the data changes the inductive bias of the estimator to prefer a model that relies more heavily on low-variance features. 
This data-dependent norm leads to good performance in-domain where such features exhibit low variation, but will result in poor performance on OOD examples where the data points are corrupted or sampled from a different distribution than that of the one observed at training time.

\subsection{Subtracting the mean does not change the estimated parameters}
\label{sec:subtracting_by_mean_is_ok}

In this section, we show that subtracting mean from each feature and target does not change the estimated parameter in the overparametrized regime. 
Recall that $X \in \mathbb{R}^{n\times d}$ are the training examples and $Y \in \mathbb{R}^{n}$ are their targets. 
Let $\mu_X \in \mathbb{R}^d$ and $\mu_Y \in \mathbb{R}$ denote the mean of the features of the training data and their targets respectively. Since $\mu_X$ is a linear combination of the rows of $X$ and row operations do not change the projection matrix, the following two linear programs share the same solution:
\begin{equation}
\label{sub1}
        \hat \zeta = \argmin \|\zeta\| \ \ s.t. \quad X\zeta = Y 
\end{equation}
\begin{equation}
\label{sub2}
    \hat \theta =  \argmin \|\theta\| \ \ s.t. \quad (X-\mu_X) \theta = Y-  \vec{1}\mu_Y
\end{equation}
We conjecture that minimizing the data-dependent norm in each neural network layer leads to reliance on low variance (frequent) features, which can result in a better in-domain generalization as these feature do not exhibit high variations.
Nonetheless, in a new domain where some (or all) of the training-domain features are missing or altered (e.g., when there is some data-agnostic corruption such as Gaussian noise), Eq.~\eqref{sub1} performs better than Eq.~\eqref{sub2} as its inductive bias is independent of the training data.  

Given such observations, how should we change the regularization such that it selects for a model that performs well both in- and out-of-domain? Inspired by this analysis, we introduce a simple, yet powerful two-step approach called Counterbalancing Teacher (CT) that combines the normalized and unnormalized copies of the same network 
for robust representation learning.

\section{Counterbalancing Teacher (CT)}
Our \emph{Counterbalancing Teacher} (CT) framework aims to learn a classifier that performs well on both in- and out-of-domain data.
Our approach consists of two steps. First, we train the teacher network to optimize the primary (classification) objective. Then, we freeze the teacher's network weights and train the student via the same primary objective while regularizing its learned representation using the frozen teacher. 
We note that both the teacher and student encoders share the same network architecture, except that the teacher's does not have any BN layers.
Given a neural network student model $S$ with one or more batch normalization layer(s), we create the teacher model by cloning the student network and removing all the batch normalization layers from the cloned network. We refer to the cloned network as teacher $T$. During the training process, the teacher and student learn different network parameters.  
\label{method}
Our approach is \textit{architecture-agnostic} and can be applied to any neural network architecture with batch normalization.\\
\textbf{Training the teacher.} The teacher is trained using the $k$-class classification objective $\mathcal{L}^{T}$. Given a data pair $(x, y)$, $f_T$ is the teacher model's unnormalized prediction (or representation), whereas $p_T$ is the normalized predictions. In our experiments, we minimize the cross-entropy loss as a supervised classification objective for the teacher:
\begin{equation}
    \mathcal{L}^{T} (x, y) = \mathcal{L}_{cls}(p_{T}, y) = - \sum_{i}^k y^i\log(p_{T}^{i}).
\end{equation}
Once the teacher is fully trained, we freeze its weights and only use it to regularize the student.
% \kristy{this subsection is unreadable; clean up}
%w
\begin{figure*}[t!]
    \centering
    \subfigure[MNIST classification results]
    {\label{fig:mnist_point}\includegraphics[width=.35\linewidth]{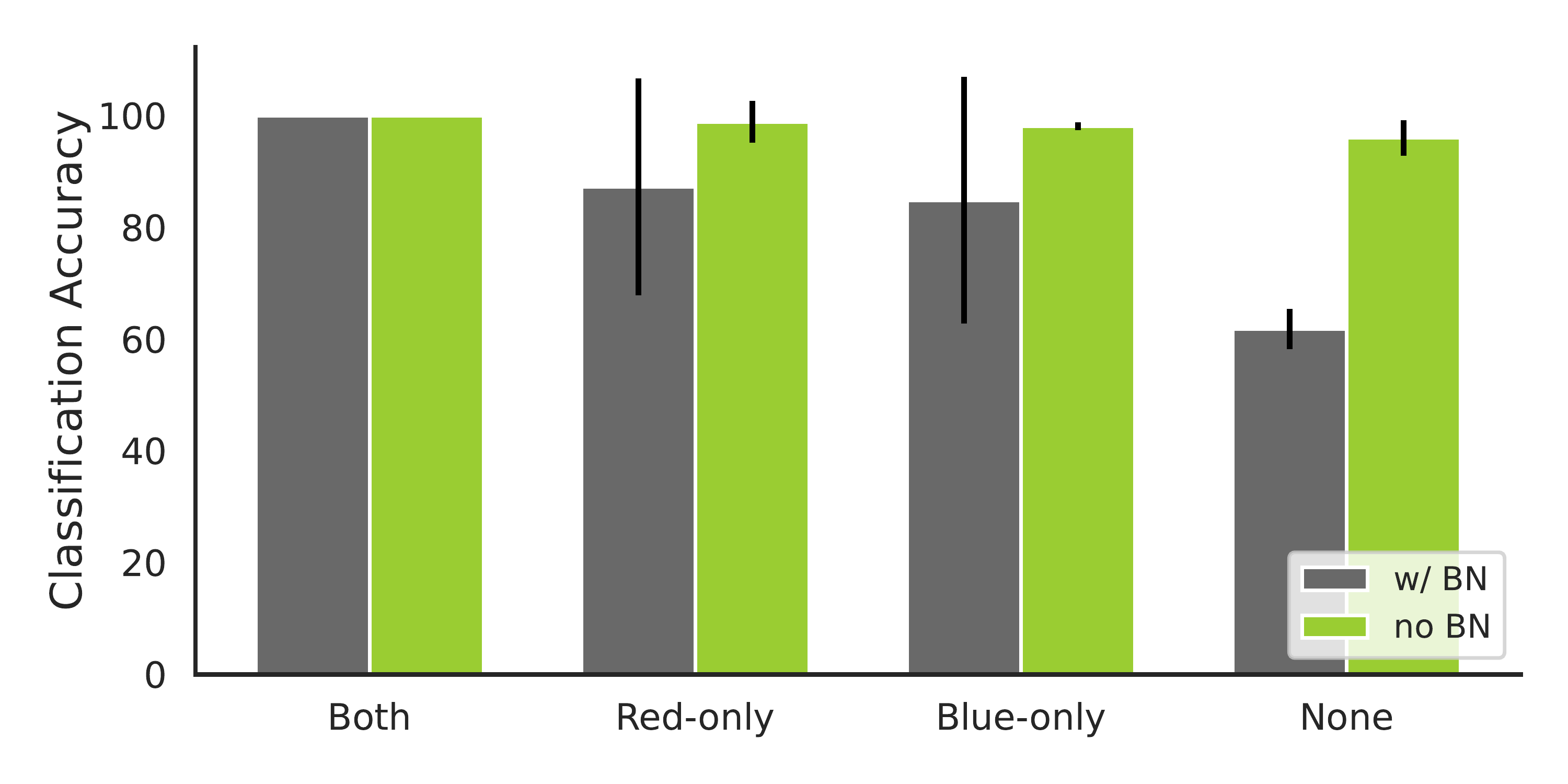}}
    \subfigure[Ablation on the ``None'' test set for $\lambda$ ]{\label{fig:mnist_lambda}\includegraphics[width=.35\linewidth]{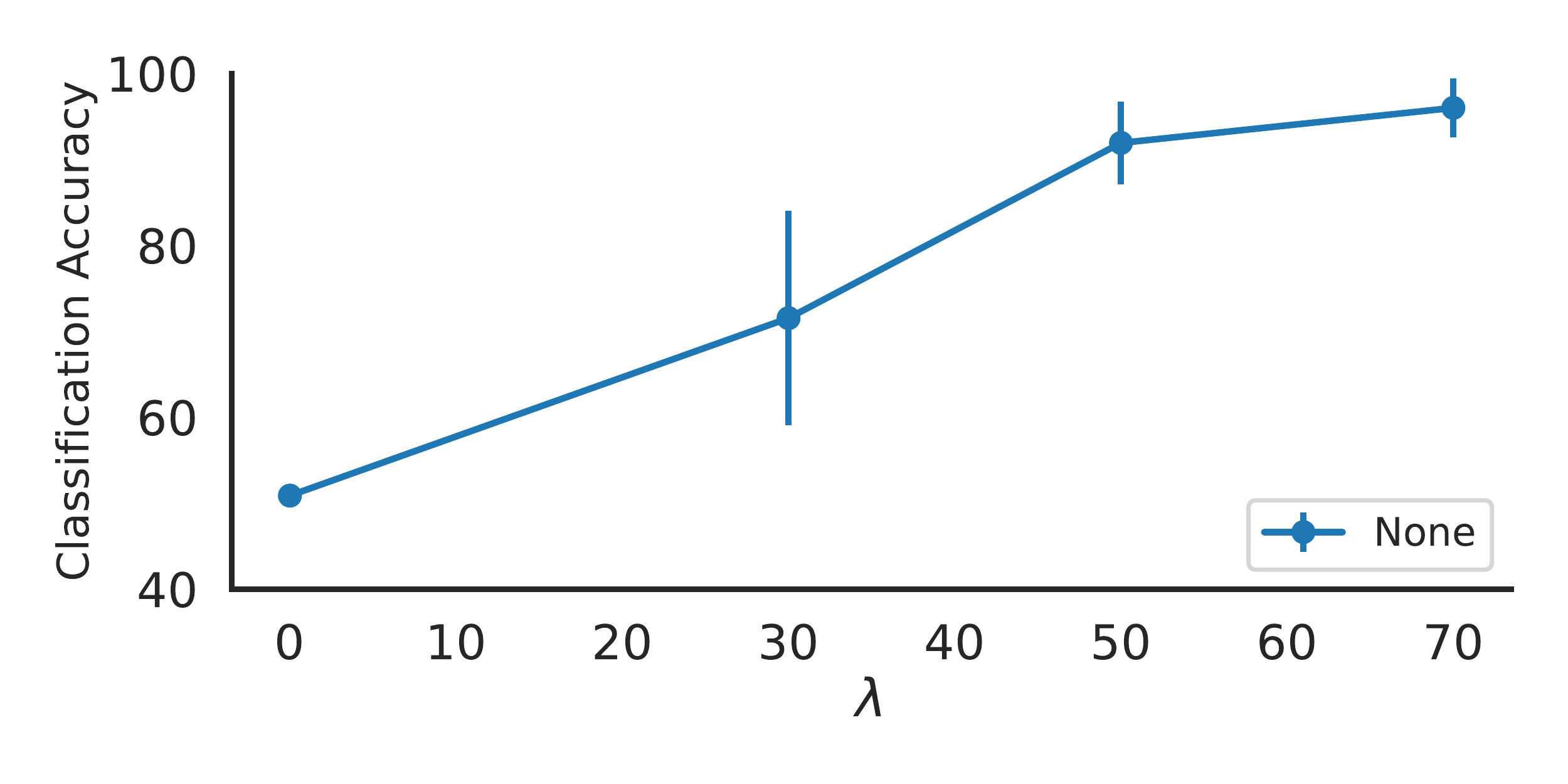}}    \subfigure[Feature importance]{\label{fig:mnist_rb}\includegraphics[width=.28\linewidth]{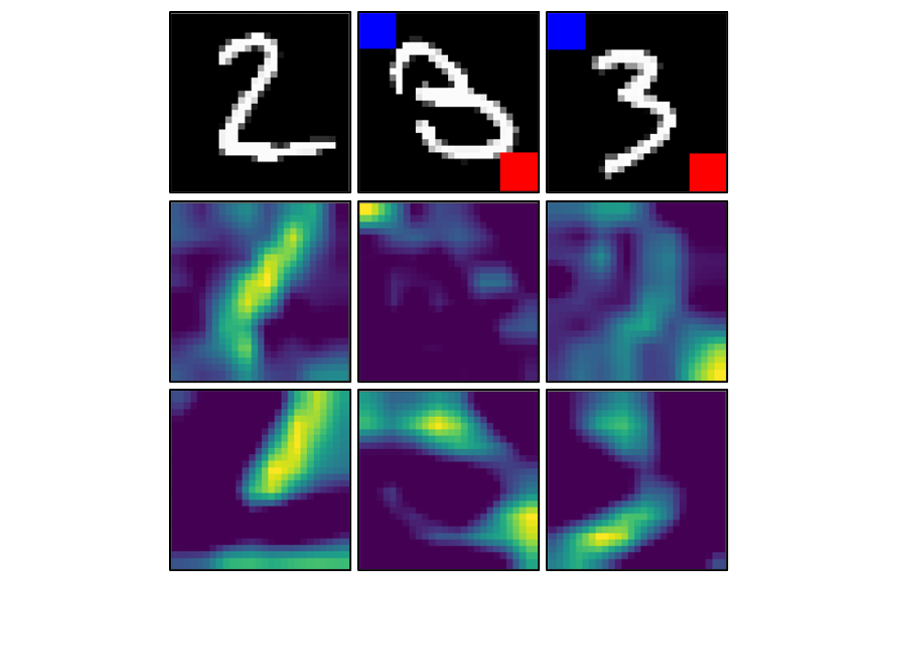}}
    \caption{(a) In the MNIST classification task, the model without BN (NoBN) achieves high classification accuracy even if the frequent training features are missing during inference. (b) Ablation results on the ``None'' test set. (c) The model with BN ($2^{nd}$ row) puts more emphasis on the red/blue squares as they are the frequent features while the NoBN model ($3^{rd}$ row) does not.}
\end{figure*}\\
\textbf{Regularizing the student for robustness with the CT.} In the second step, we start training the student model $S$ by minimizing the cross-entropy loss while regularizing it using the frozen teacher. Drawing inspiration from~\cite{huang2017arbitrary}, we propose the following objective to regularize the representations of the student  $f_{S}$:
\begin{equation}
    \mathcal{L}_{ct}\left (f_{S}, f_{T}  \right ) = \frac{1}{h}\sum_{i=1}^{h}\left (f_{T}^{i} - f_{S}^{i}   \right )^2
    \label{eq3}
\end{equation}

where $f_{S} \in \mathbb{R}^h$ and $f_{T} \in \mathbb{R}^h$ refer to the $h$ dimensional representations learned by the student $S$ and teacher $T$, respectively. Therefore, the final objective function used to train the student becomes:
\begin{equation}
        \mathcal{L}^S(x, y) = \mathcal{L}_{cls}(p_{S}, y) + \lambda \mathcal{L}_{ct}(f_T, f_S)
    \label{eq:loss}
\end{equation} 

where $\lambda > 0$ is a hyperparameter controlling the amount of regularization. 

Our approach closely follows ``offline distillation''~\citep{hinton2015distilling}, where the knowledge from a pre-trained teacher network is distilled into a student network.
We observed empirically that the two-step training approach outperformed joint training of the student and teacher simultaneously.

\section{Experimental Results}

\label{experiments}
In this section, we are interested in empirically investigating the following questions:
\begin{enumerate}
    \item Do batch and other normalization layers indeed incentivize neural networks to learn low-variance representations? 
    % in neural networks indeed lead to learning frequent low variance features? 
    \item Would a trained model with normalization fail if the dominant features are missing or corrupted, and how would an unnormalized network behave in the same scenario?
    \item Does CT improve in- and out-of-domain accuracies as well as low corruption errors across different architectures and data modalities?
\end{enumerate}
% \kristy{todo}

\subsection{Batch normalization is a cause of failure when frequent low variance input features are missing}

We warm up with a synthetic experiment to test the hypothesis that batch normalization leads to learning low-variance features. 
We design a binary classification problem by selecting two digits of ``2'' ($y=0$) and ``3'' ($y=1$) from the MNIST dataset, and add a red and a blue square to all samples of class ``3'' (Figure~\ref{fig:mnist_rb} first row). Since these nuisance features are only added to one class, they represent the low-variance features. We add Gaussian noise to samples before adding the squares to ensure that the coloured squares are the only low variance features. 
% \kristy{talk about the different test sets here}

We examine this hypothesis both qualitatively (visualizing important features for the trained classifiers using GradCAM~\citep{selvaraju2017grad}) and quantitatively by evaluating on multiple test sets: 1) \texttt{Both}: similar to training samples, digit ``3'' includes both the red and blue squares, 2) \texttt{Red-only}: digit ``3'' includes only red square, 3) \texttt{Blue-only}: digit ``3'' includes only blue square, and 4) \texttt{None}: contains no square. 

Ideally, the trained classifier should not rely \textit{only} on the dominant features for the prediction task. 
However, as shown in the second row of Figure~\ref{fig:mnist_rb}, the network trained with batch normalization quickly picks up those low-variance clues, which in this case is either the red or the blue square. 
However, the same network without normalization takes into account other features as well. We verify this by quantifying classification accuracy on the different test sets. As demonstrated in Figure~\ref{fig:mnist_point}, since the model relies mostly on the colored squares, it drastically fails when those features are missing at test time (the `\texttt{None} test set) which is not the case for the network without normalization.

\begin{figure*}[t!]
    \centering
    \subfigure[ResNext]{\includegraphics[width=.49\linewidth]{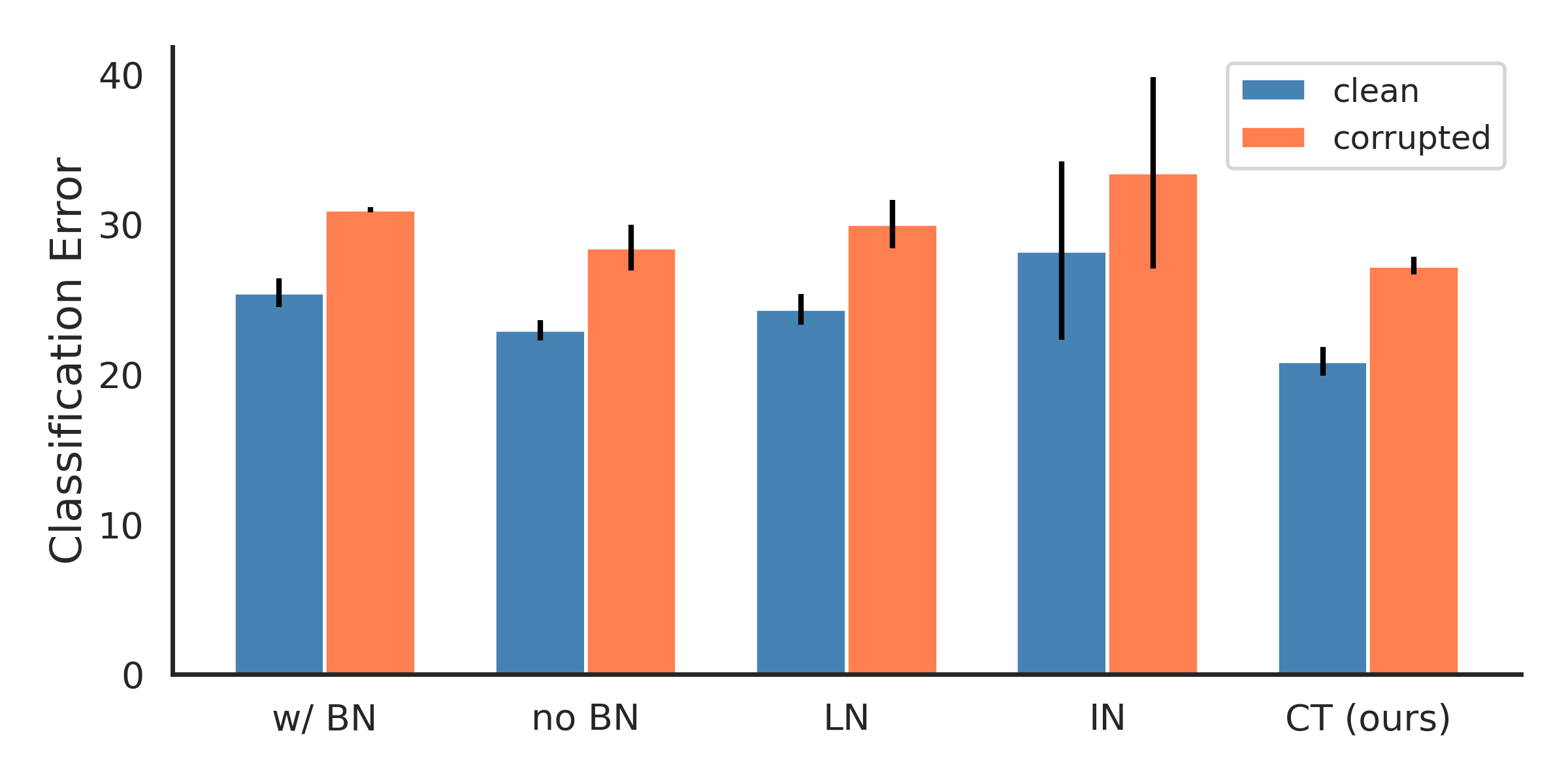}}
    \subfigure[WideResNet]{\includegraphics[width=.49\linewidth]{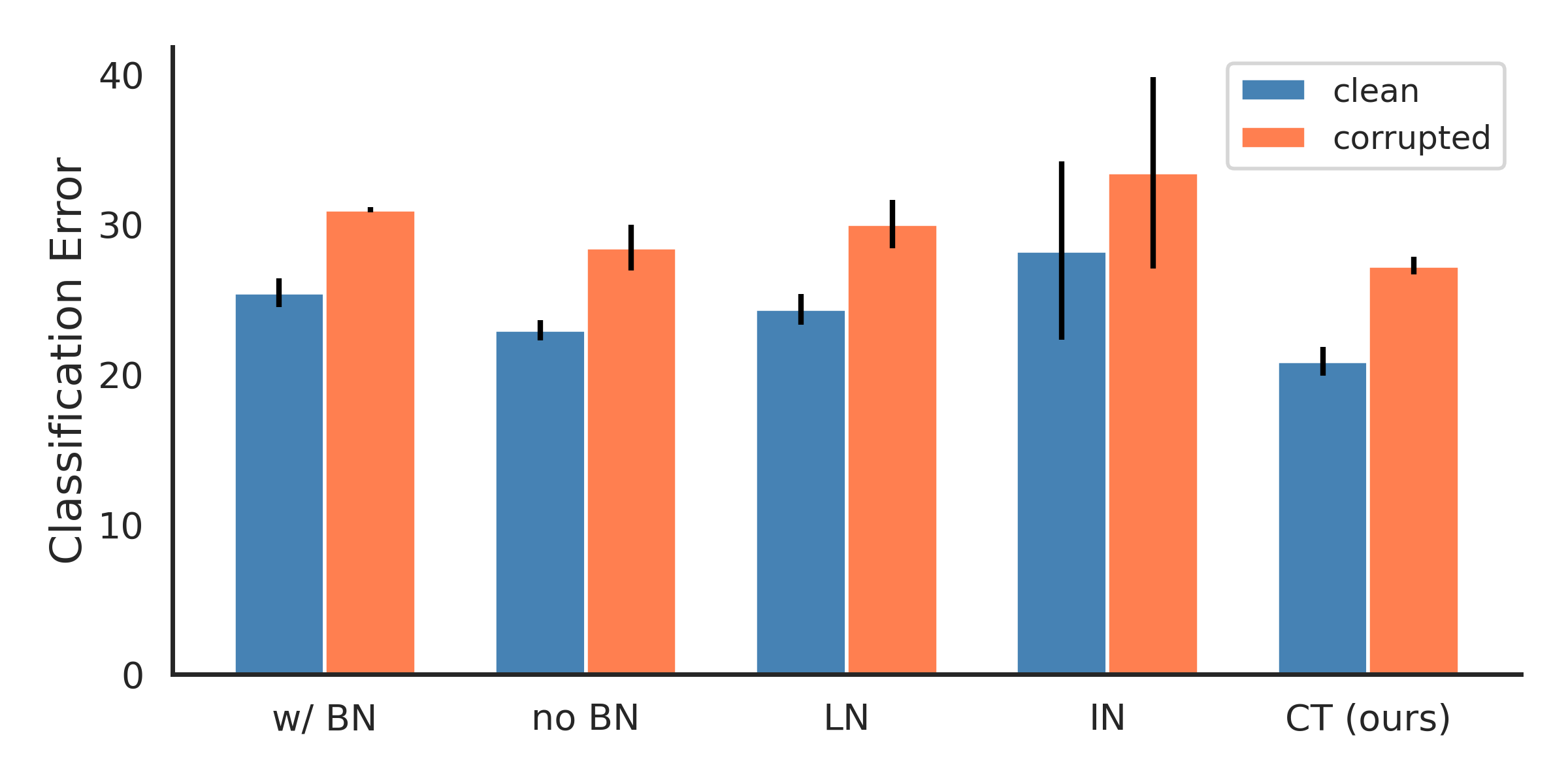}}
    \caption{CIFAR-10-C error. For both the models, CT achieves lower corruption errors compare to normalizing methods.}%\kristy{be more descriptive}}
    \label{fig:cifar10_noaug}
\end{figure*}

\subsection{Batch normalization is a cause of failure when input features are corrupted}

In this experiment, we focus on a dataset corruption scenario where models are trained with clean data while the test data is corrupted by commonly used perturbations in the literature~\cite{hendrycks2019augmix}. 
We test the hypothesis that our model should learn representations that lead to a better performance than normalized models when corruptions are applied to the data at test time.
Here, we use the WideResNet 40-2, AllConvNet, and ResNext architectures and CIFAR-10-C dataset to be consistent with prior works~\cite{hendrycks2019augmix}. 
For simplicity and to directly study the effect of normalization under data shifts we do not use dropout and data augmentations in this experiment.% \kristy{do the other methods?}. 
We further replace batch normalization with other common normalization techniques such as layer normalization (LN)~\citep{ba2016layer} and instance normalization (IN)~\citep{ulyanov2016instance}. 

As shown in Figure~\ref{fig:cifar10_noaug}, the model with no normalization (NN) outperforms all the three types of normalization when evaluated on corrupted data. 
As expected on clean data, all three approaches perform favorably. With our CT model, corruption errors improves significantly on WideResNet and ResNext, by $\sim4\%$ and $\sim3\%$, respectively. 

\subsection{Regularizing a batch normalized model using CT leads to a higher robustness}
Here we compare our CT model with recent approaches on robustness, multi-domain generalization, and corrupted point cloud classification using different data sets and models. For experimental setting, refer to Appendix~\ref{exp_details}. 

\paragraph{Robustness to input data corruptions.}
%\kristy{todo; fix writing}
We compare the robustness of our CT model to two main groups of the recent approaches on robustness: 1) covariate shift adaptation~\citep{benz2021revisiting,schneider2020improving} and 2) data augmentation-based methods~\citep{hendrycks2019benchmarking} on CIFAR-10-C and CIFAR-100-C~\citep{hendrycks2019benchmarking}. 
%\kristy{ImageNet?}
Covariate shift adaptation approaches are designed with the assumption that there exist some samples from (or close to test distribution) which can be used to calibrate the parameters of the batch normalization layers. 
However, we show that these approaches work only if BN statistics are adapted to a particular corruption. 
This is an unrealistic assumption in practice, as we often do not have prior knowledge of corruption types that may or may not occur at test time. 
Additionally, we show adapting batch norm statistics to a single or a few distortion types can often fail to generalize to unseen distortions~\citep{geirhos2018generalisation,vasiljevic2016examining} which is also the case for data augmentation-based methods.

% \looseness=-1
\paragraph{CT vs. covariate shift adaptation methods for robustness.} 
We consider three scenarios for adaptation of batch norm statistics; Let $C = \{c_i\ |\ i \in Z\}$ be the set of corruptions that can be applied to the input of the network (we assume that the corruption severity level~\citep{hendrycks2019benchmarking} is 3 in all scenarios):
\begin{enumerate}[label=(\roman*)]

\itemsep0em 
\item \textbf{Adapt-one-test-one}: BN statistics are adapted to a batch of samples from each corruption type $c_i$ at a time, and classification error is computed on the same $c_i$. This is repeated for each corruption type and classification error is averaged over each type \citep{benz2021revisiting,schneider2020improving}. However, it is not trivial to differentiate samples based on corruption type and adapt each separately at test time.
\item \textbf{Adapt-one-test-all}: BN statistics are adapted to a batch of samples from a randomly selected $c_i$, and the classification error is averaged over $C$. This experiment reveals the caveats of adaptation on a single corruption, where model fails to generalize to other types of corruptions.
\item \textbf{Adapt-all-test-all}: BN statistics are adapted to a batch of samples from a \textit{random combination} of samples from $C$. The goal is to aid adaptation of statistics by looking at samples from ``all'' corruption types. The classification error is then averaged over $C$. This scenario is more realistic than (I) and (II), but still requires a batch of samples from the target domain. 
\end{enumerate}

\begin{figure*}[t!]
%   \captionsetup{justification=centering}%
  \centering
    % \vskip-\intextsep
    \includegraphics[width=\textwidth]{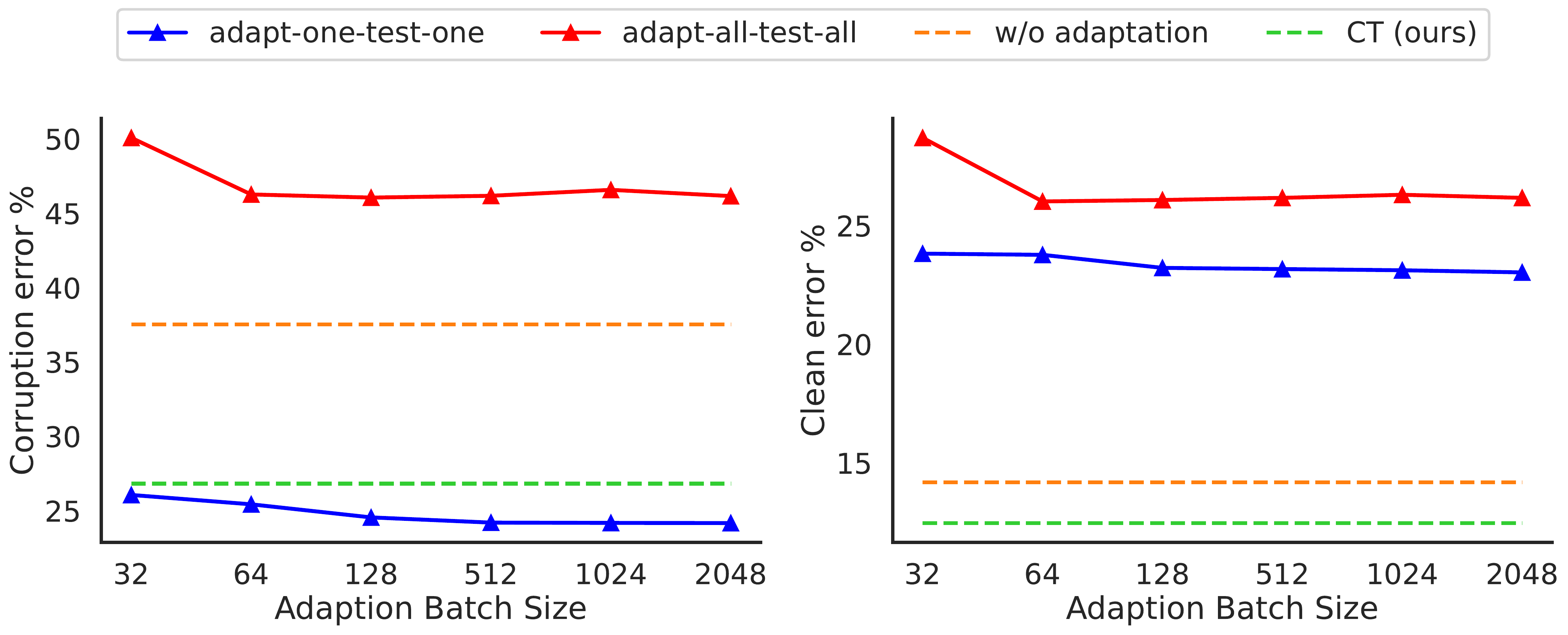}
  \caption{Corruption (left) and clean (right) errors by increasing adaptation batch size on CIFAR-10-C. Covariate shift adaptation-based methods fail in a more realistic setting (\texttt{adapt-all-test-all}) i.e. updating the batch norm statistics using randomly selected batches of samples from the target domain, while CT performs significantly better without requiring the target domain knowledge/statistics.}
    % \vskip-\intextsep
    \label{fig:bn_adapt_comp}
\end{figure*}

In Figure~\ref{fig:bn_adapt_comp}, we shed light on a more realistic scenario of batch norm statistics adaptation. We assume access to some extra samples, but not their corruption types (i.e., \texttt{adapt-all-test-all}). Surprisingly, we regardless of batch size, such methods perform significantly worse on both clean and corruption errors compared to our CT method on CIFAR-10-C with WideResNet model. Further analysis on how adaptation-based methods cause generalization failures are included in Table~\ref{tab:adapt_one_test_all}.

\paragraph{Comparing CT to data augmentation-based methods.}~\cite{hendrycks2019augmix} has studied the effectiveness of different data augmentation techniques on making a model robust to common data corruptions. They proposed to create a weighted mixture of input samples using multiple different augmentation types (AugMix) and showed this method can outperform other augmentation-based methods on robustness to common corruptions. Since our CT model is inherently data augmentation-independent, \textit{without any augmentation}, it outperforms (by achieving 25.7\% mCE while the original mCE is 39.6\%) several recent and widely used augmentation-based approaches such as Standard~\citep{hendrycks2019augmix}, Cutout~\citep{devries2017improved}, CutMix~\citep{yun2019cutmix}, and Adversarial Training~\citep{madry2017towards} on CIFAR-10-C with WideResNet. When CT is used with a few simple data augmentations (AutoAugment~\citep{cubuk2018autoaugment} individual operations, but randomly), as shown in Table~\ref{tab:data_augs}, it outperforms most of the data augmentation-based methods including the recent and advanced ones such as Mixup~\citep{zhang2017mixup} and  AutoAugment~\citep{cubuk2018autoaugment}, while it achieves comparable results to AugMix without the Jensen-Shannon divergence (JSD) loss on CIFAR-10-C and CIFAR-100-C datasets with different networks.

{
\setlength{\tabcolsep}{1.7pt}
\begin{table*}
\centering
\resizebox{\textwidth}{!}{%
\begin{tabular}{llcccccccc}
\cmidrule[0.08em]{3-10}
                             &  & Standard & Cutout & Mixup & CutMix & AutoAug & AdvT & AugMix  & CT (ours) \\ \midrule
\multirow{2}{*}{CIFAR-10-C}  & AllConvNet & 30.8     & 32.9   & 24.6  & 31.3   & 29.2        & 28.1     & 15.0    & \textbf{14.0$\pm$0.55}     \\
                             & WideResNet & 26.9     & 26.8   & 22.3  & 27.1   & 23.9        & 26.2     & \textbf{11.2}   & 11.6$\pm$0.15     \\ \midrule
\multirow{2}{*}{CIFAR-100-C} & AllConvNet & 56.4     & 56.8   & 53.4  & 56.0   & 55.1        & 56.0     & 42.7    & \textbf{39.8$\pm$0.06}      \\
                             & WideResNet & 53.3     & 53.5   & 50.4  & 52.9   & 49.6        & 55.1     & 35.9    & \textbf{33.7$\pm$0.12}     \\ \bottomrule
\end{tabular}}
\caption{CIFAR-10-C and CIFAR-100-C mean corruption error (mCE) compared to common data augmentation techniques. AdvT, and AutoAug refer to adversarial training, Auto Augment, respectively. Our CT approach outperforms six out of seven methods while it achieves comparable results to AugMix which leverages complex augmentations. The clean error on CIFAR-10 for CT on AllConvNet is $8.2 \pm 0.28$ and WideResNet is $7.0 \pm 0.06$. The clean error on CIFAR-100 AllConvNet is $31.46 \pm 0.20$ and WideResNet is $26.39 \pm 0.11$. This demonstrates that CT has good ID performance.}
\label{tab:data_augs}
\end{table*}}

\paragraph{CT's performance in domain generalization.} In this experiment, we evaluate CT on domain generalization using the VLCS benchmark ~\citep{torralba2011unbiased}. VLCS consists of images from five object categories shared by the PASCAL VOC 2007, LabelMe, Caltech, and Sun datasets, which are considered to be four separate domains. We follow the standard evaluation strategy used in~\citep{carlucci2019domain}, where we partition each domain into a train (70\%) and test set (30\%) by random selection from the overall dataset. We use ResNet-18 as the backbone to make a fair comparison with the state-of-the-art. As summarized in Table~\ref{tab:ood}, CT outperforms the state-of-the-art on 3 out of 4 domains and by 1.83\% on average.

\begin{table*}[h!]
\setlength{\tabcolsep}{5pt}
\renewcommand{\arraystretch}{1}
\centering
\resizebox{\textwidth}{!}{%
\begin{tabular}{lccccc}
\toprule
\textbf{Method}        & Caltech & LabelMe & Pascal & Sun   & Average  \\ \midrule
DeepC~\citep{li2018deep}         & 87.47   & 62.06   & 64.93  & 61.51 & 68.89 \\
CIDDG~\citep{li2018deep}         & 88.83   & 63.06   & 64.38  & 62.10 & 69.59 \\
CCSA~\citep{motiian2017unified}          & 92.30   & 62.10   & 67.10  & 59.10 & 70.15 \\
SLRC~\citep{ding2017deep}          & 92.76   & 62.34   & 65.25  & 63.54 & 70.15 \\
TF~\citep{li2017deeper}            & 93.63   & 63.49   & 69.99  & 61.32 & 72.11 \\
MMD-AAE~\citep{li2018domain}       & 94.40   & 62.60   & 67.70  & 64.40 & 72.28 \\
D-SAM~\citep{d2018domain}         & 91.75   & 57.95   & 58.59  & 60.84 & 67.03 \\
%JiGen~\citep{carlucci2019domain}         & 96.93   & 60.90   & 70.62  & 64.30 & 73.19 \\
Shape Bias~\citep{asadi2019towards} & 98.11   & 63.61   & \textbf{74.33}  & 67.11 & 75.79 \\ 
CT (ours) & \textbf{99.21}   & \textbf{65.87}  & 74.10  &\textbf{71.20} & \textbf{77.60} \\ 
\bottomrule
\end{tabular}}
\caption{Multi-source domain generalization accuracy (\%) on the VLCS dataset with ResNet-18 as the base network for classification. All reported numbers are averaged over three runs.}
\label{tab:ood}
\end{table*}

\paragraph{CT's classification performance on corrupted 3D point cloud data.} In this experiment, we use the RobustPointSet  dataset~\cite{taghanaki2020robustpointset} which is created for analysis of point classifiers in terms of robustness to 3D corruptions. We follow the same training-domain validation setting as in~\cite{taghanaki2020robustpointset}. We train each model without data augmentation on the clean training set and select the best performing checkpoint on the clean validation set for each method. We then test the models on the six distorted unseen test sets. This experiment shows the vulnerability of the models trained on original data to unseen input transformations. As shown in Table~\ref{tab:robustpointset}, PointNet-CT significantly improves classifcation accuracy on most of the shifted test sets, and 1.87\% on average compared to original PointNet.

\begin{table*}[h!] 
\centering 
\setlength{\tabcolsep}{2pt}
\resizebox{\textwidth}{!}{%
\begin{tabular}{lcccccccc}
\toprule
                                Method             & Original & Noise & Translation & MissingPart & Sparse & Rotation & Occlusion & Avg. \\ \midrule
                                    %PointNet++ (M)   & 91.27    & 5.73  & 91.31       & 53.69        & 6.65   & 13.02    & 64.18     & 46.55   \\
                                    %DGCNN             & 92.52    & 57.56 & 91.99       & 85.40        & 9.34   & 13.43    & 78.72     & 61.28   \\
                                    %PointMask         & 88.53    & 73.14 & 78.20       & 81.48        & 58.23  & 8.02     & 39.18     & 60.97   \\
                                    %DensePoint         & 90.96    & 53.28 & 90.72       & 84.49        & 15.52  & 12.76    & 67.67     & 59.40   \\
                                    %PointCNN         & 87.66    & 45.55 & 82.85       & 77.60        & 4.01    & 11.50    & 59.50     & 52.67   \\
                                    %PointConv        & 91.15    & 20.71 & 90.99       & 84.09        & 8.65  & 12.38     & 45.83     & 50.54 \\
                                    %R-ShapeCNN & 91.77    & 48.06 & 91.29       & 85.98        & 23.18  & 11.51    & 75.61     & 61.06   \\ 
                                    PointNet            & 89.06    & 74.72 & 79.66  & 81.52   &60.53  & \textbf{8.83}     & \textbf{39.47}     & 61.97 \\
                                    PointNet-noBN     & 87.99    & 78.65 & 79.02  & 76.82   &72.61  & 6.89     & 35.13     & 62.44 \\
                                    PointNet-CT (ours)  & \textbf{89.83}    & \textbf{78.12} & \textbf{82.25}  & \textbf{82.58}   &\textbf{68.40}  & 7.62     & 38.10     &
                                    \textbf{63.84} \\ %\midrule
                                    % PointNet++ (M) & \textbf{91.27} &	5.73 &	\textbf{91.31} &	\textbf{53.69} &	6.65 &	\textbf{13.02} &	64.18 &	46.55 \\
                                    % PointNet++ (M)-NoNorm & 85.79 & 72.25 & 78.2 & 76.26 & 77.53 & 7.27 &  32.79 & 61.44\\
                                    % PointNet++ (M)-CT (ours) & 90.31 & \textbf{20.44} & 90.19 & 41.43 & \textbf{7.98} & 11.86 & \textbf{68.03} & \textbf{46.98}\\
                                    
                                    \bottomrule
\end{tabular}}
\caption{Robust classification accuracy on RobustPointSet. The \texttt{Noise} column for example shows the result of training on the \texttt{Original} train set and testing with the \texttt{Noise} test set. When we train PointNet using our CT method its performance significantly improves on average and particularly on \texttt{Noise}, \texttt{Translation}, \texttt{MissingPart}, and \texttt{Sparse} test sets. } 
\label{tab:robustpointset}
\end{table*}

\begin{table*}[h!]
\setlength{\tabcolsep}{4pt}
\renewcommand{\arraystretch}{1.1}
\resizebox{\textwidth}{!}{%
\begin{tabular}{lccc|ccc|ccc|cc}
\toprule
          & \multicolumn{3}{c|}{Avg. Cal. Err.}              & \multicolumn{3}{c|}{MA Adv. Cal. Err.}           & \multicolumn{3}{c|}{RMS Adv. Cal. Err.}          & \multicolumn{2}{c}{Scoring Rules} \\
          & RMS            & MA             & MCA   & GS=0.11        & GS=0.56        & GS=1.00        & GS=0.11        & GS=0.56        & GS=1.00        & Sharpness       & CRPS            \\ \hline
Baseline  & 0.378          & 0.341          & 0.344          & 0.358          & 0.345          & 0.341          & 0.392          & 0.383          & 0.378          & 0.014           & 0.481           \\
CT (ours) & \textbf{0.142} & \textbf{0.116} & \textbf{0.117} & \textbf{0.125} & \textbf{0.118} & \textbf{0.116} & \textbf{0.153} & \textbf{0.146} & \textbf{0.142} & \textbf{0.064}  & \textbf{0.041}  \\\bottomrule
\end{tabular}}
\caption{Uncertainty analysis results, comparing the baseline (presented in Fig. \ref{fig:mnist_point} as \textit{wBN}) against our CT method. Our method significantly outperforms the normalized model. Acronyms used in the table: \textbf{RMS}: Root Mean Squared, \textbf{MA}: Mean Absolute, \textbf{MCA}: Miscalibration Area, \textbf{GS}: Group Size.}

\label{tab:uncertainty_analysis}
\end{table*}

\paragraph{Uncertainty analysis of CT.} We analyze the predictive uncertainty of our CT and compare it with the baseline (wBN). 
To do so, we calculated the Calibration Error (Cal. Err.)~\cite{guo2017calibration}, Adversarial Calibration Error (Adv. Cal. Err.), and Scoring Rules~\cite{gneiting2007probabilistic,deisenroth2020mathematics} for both the baseline and our method and reported in the Table~\ref{tab:uncertainty_analysis}. For more detailed explanation of the metrics we refer the reader to Appendix~\ref{uncertainty-app}.

\section{Conclusion}
\label{conclusion}
In this work, we investigated the effect of BN on model robustness. We provided a theoretical justification in the overparameterized linear regime for how BN compromises OOD generalization performance by encouraging the model to rely on highly predictive, low-variance input features. Then, we proposed a robust representation learning framework---Counterbalancing Teacher (CT)--- which leverages a frozen copy of the same model without BN as a teacher to regularize the features learned by a student network to improve generalization. Empirically, we demonstrated that our method's learned representations are robust to common corruptions on a suite of domain generalization tasks. Similar to knowledge distillation, a limitation of our CT method can be training an extra encoder as a regularizer. However, since the encoder is discarded after training, the inference time for CT remains the same to that of the original model.

\bibliography{outline}
\bibliographystyle{apalike}

\newpage
\appendix
\section*{Appendix}
\renewcommand{\thesubsection}{\Alph{subsection}}
\label{appendix}

\subsection{Additional experimental details}
\label{exp_details}

We implement our CT variants using Tensorflow~\cite{abadi2016tensorflow}.

\subsubsection{MNIST missing features experiment}
We trained the MNIST models (architecture below) with SGD optimizer (with Tensorflow's default parameters) and batch size of 128 for 100 epochs. For the NoBN model, we simply removed the BN layers and trained all models without any data augmentation.

\begin{table*}[!ht]
\centering
\caption{MNIST small CNN architecture with batch normalization layers. Batch normalization layers (in gray) are omitted in the Teacher network.}
\begin{tabular}{ll}
\toprule
\textbf{\#} & \textbf{Layer} \\
\midrule
1 & Conv2D (in=d, out=16, stride=1) \\
2 & \textcolor{gray}{BatchNorm} \\
3 & ReLU \\
4 & Conv2D (in=16, out=16, stride=1) \\
5 & \textcolor{gray}{BatchNorm} \\
6 & ReLU \\
7 & MaxPool2D (stride=2) \\
8 & Conv2D (in=16, out=32, stride=1) \\
9 & \textcolor{gray}{BatchNorm} \\
10 & ReLU \\
11 & Conv2D (in=32, out=32, stride=1) \\
12 & \textcolor{gray}{BatchNorm} \\
13 & ReLU \\
14 & MaxPool2D (stride=2) \\
15 & Dense (nodes=256) \\
16 & \textcolor{gray}{BatchNorm} \\
17 & ReLU \\
18 & Dense (nodes=c) \\
19 & Softmax \\
\bottomrule
\end{tabular}

\end{table*}

\subsubsection{Comparing CT to other methods on common input corruption}
For both CIFAR-10-C and CIFAR-100-C we train $\mathcal{G}_\theta$ using Adam optimizer with learning rate of 0.0001 and batch size of 64. In the second step, we train $\mathcal{F}_\zeta$ using SGD with Nesterov, initial learning rate of 0.1 and decaying to 0.00001 using cosine scheduler for 300 epochs. 

% To be fair when comparing to other extensive data augmentation techniques such as AugMix, AutoAugment and CutMix, we applied simple augmentations (AutoAugment operations, but randomly) with RGB channels shuffle and randomly masking up to 50\% of the input pixels. 

\subsubsection{Applying CT to point clouds}
We trained both $\mathcal{G}_\theta$ and $\mathcal{F}_\zeta$ with Adam optimizer and learning rate of 0.001 (divided by 10 at epochs 50 and 75) and batch size of 32 for 500 epochs. 

\subsubsection{Applying CT to domain generalization}
We trained CT with SGD optimizer with learning rate of 0.001 and momentum of 0.9. We set batch size to 32 and image size to 224x224, and train for 500 epochs.

\subsubsection{Self-supervised baselines}
We train the self-supervised methods with batch size 256 for 1000 epochs while maintaining the original hyper-parameters. We fine-tune all self-supervised methods for 350 epochs with Adam~\cite{kingma2014adam} and with starting learning rate of 0.001 with a cosine decay, and $l_2$ coefficient set to 0.0005 to avoid overfitting.

\subsection{Additional experimental results}

\subsubsection{Comparing the robustness of CT to self-supervised methods on common input corruptions.}

In this experiment, we evaluate how recent self-supervised and contrastive learning methods perform on unseen input corruptions: SimCLR~\citep{chen2020simple}, BYOL~\citep{grill2020bootstrap}, and Barlow Twins~\citep{zbontar2021barlow}. As reported in Table~\ref{tab:self_supervised}, our method outperforms all self-supervised methods (with fine-tuned encoders) based on the mean corruption error for both CIFAR-10-C and CIFAR-100-C by a large margin. For all the methods in Table~\ref{tab:self_supervised} we use the same individual augmentations (same operations as in AutoAugment) as in other experiments. See appendix~\ref{ss_linear} for linear evaluation results. Concretely, we note that these self-supervised representation learning methods proceed in two stages: pre-training an encoder via self-supervision; and learning a linear classification head on top of the pre-trained encodings. To make it a fair comparison, our modification was to convert this two-stage process into a single joint-training procedure, by using an objective that combines these two terms: (1) a contrastive loss term that uses label information to select positive and negative examples for training the encoder (similar to Supervised Contrastive Learning by~\citep{khosla2020supervised}); and (2) a cross-entropy loss term for the downstream classification task. In this way, both the encoder and classifier benefit from labeled supervision. 

\begin{table*}[h]
    \centering
\begin{tabular}{llccc|c}
\cmidrule[0.08em]{3-6}
                & & SimCLR  & BYOL  & Barlow Twins &CT (ours) \\ \midrule
\multirow{2}{*}{CIFAR-10-C}  & AllConvNet & 30.87 &  34.12 & 34.35 & \textbf{14.0$\pm$0.55}     \\
                             & WideResNet & 23.12 &  26.04 & 25.80  & \textbf{11.6$\pm$0.15}     \\ \midrule
\multirow{2}{*}{CIFAR-100-C} & AllConvNet & 57.51 &  63.76 & 60.07  & \textbf{39.8$\pm$0.06}      \\
                             & WideResNet & 57.59 & 56.40  & 55.88 & \textbf{33.7$\pm$0.12}     \\ \bottomrule
\end{tabular} 
\caption{CT's mean corruption error (mCE) compared to common self-supervised and contrastive learning methods on CIFAR-10-C and CIFAR-100-C. All SSL models are first pre-trained (self-supervised), then fine-tuned (supervised).}
    \label{tab:self_supervised}
\end{table*}

\subsubsection{Linear evaluation of self-supervised models}
\label{ss_linear}
Self-supervised linear evaluation results where encoders are frozen and only classification layer is trained.
{ 
% % \setlength{\tabcolsep}{4pt}
\begin{table}[ht!]
\centering
\begin{tabular}{llccc|c}
\cmidrule[0.08em]{3-6}
                & & SimCLR   & BYOL  & Barlow Twins &CT (ours) \\ \midrule
\multirow{2}{*}{CIFAR-10-C}  & AllConvNet & 58.84 & 65.47  & 58.20 & 16.9     \\
                             & WideResNet & 58.18 &  53.32 & 59.75 &13.9     \\ \midrule
\multirow{2}{*}{CIFAR-100-C} & AllConvNet & 82.91 &  91.20 & 86.16 &42.6      \\
                             & WideResNet & 84.08 & 83.38  & 85.61 & 43.1     \\ \bottomrule
\end{tabular} 

\caption{CT's mean corruption error (mCE) compared to common self-supervised and contrastive learning methods on CIFAR-10-C and CIFAR-100-C. Encoders are frozen and only classification layer is trained (linear evaluation).}
\label{tab:self_supervised_linear}
\end{table}}

\subsubsection{Batch-norm adaptation additional results}
Batch norm adaptation-based results for robustness to common input corruptions on CIFAR-10-C.

{\renewcommand{\arraystretch}{1}
\begin{table*}[h!]
    \centering
    \setlength{\tabcolsep}{3pt}
\resizebox{\textwidth}{!}{%
        \begin{tabular}{l|ccc|cccccccc}
        \toprule
            CRP & BS & CLN-E & mCE & Gauss. & Shot & Impulse & Defocus & Glass & Motion & Snow & Compress \\
            \midrule
None & - & 14.2 & 37.54 & 56.26 & 47.93 & 49.28 & 29.54 & 56.44 & 38.51 & 32.35 & 32.71 \\
\midrule
a-o-t-o & 32 & 23.83 & 26.07 & 37.72 & 35.03 & 34.82 & 16.84 & 40.03 & 24.86 & 27.86 & 31.1 \\
\midrule
Impulse & 32 & 29.04 & 49.88 & 45.92 & 41 & 39.52 & 56.97 & 60.46 & 65.79 & 40.31 & 45.51 \\
Impulse & 16 & 33.06 & 55.02 & 51.39 & 45.73 & 43.62 & 62.27 & 66.23 & 71.03 & 45.62 & 51.12 \\
Impulse & 2 & 43.7 & 61.47 & 62.31 & 58.16 & 58.33 & 62.68 & 73.32 & 68.68 & 56.95 & 59.56 \\
Defocus & 32 & 20.39 & 41.98 & 70.47 & 62.49 & 59.07 & 21.54 & 65.94 & 35.58 & 41.16 & 38.9 \\
Defocus & 16 & 21.11 & 44.43 & 72.24 & 64.88 & 58.77 & 24.53 & 69.55 & 38.36 & 43 & 41.64 \\
Defocus & 2 & 44.47 & 62.1 & 74.39 & 70.69 & 68.91 & 51.03 & 74.98 & 54.16 & 64.18 & 68.92 \\
            \bottomrule
        \end{tabular}}
    \caption{Changes in clean error (CLN-E), mean corruption error (mCE), and corruption error with respect to changes in adaptation batch size (BS). In this table, the model's batch normalization layers are only adapted to one corruption (CRP) based on the method presented in \cite{benz2021revisiting}. This table shows an example of adapt-one-test-all scenario. All numbers are reported with the WideResNet40-2 model. Adaptation of batch norm statistics to Impulse noise, when observed test-time data is adequate, will improve the robustness to similar corruption types (Gaussian, Shot and Impulse noise). However it significantly reduces generalization to other corruption types. Top row shows original model performance without adaptation. Second row shows \texttt{adapt-one-test-one} results.}
    \label{tab:adapt_one_test_all}
\end{table*}
}

\subsubsection{Predictive performance of MNIST dataset variables.}
\begin{figure}[t!]
%   \captionsetup{justification=centering}%
  \centering
    \includegraphics[width=0.8\textwidth]{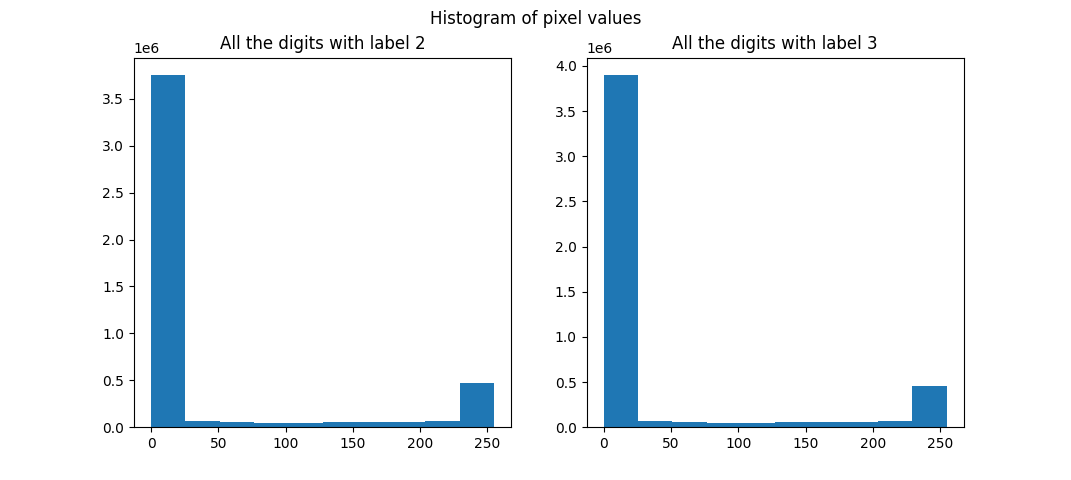}
  \caption{Histogram plots of class `2' vs. `3'.}
    \label{fig:mnist_hist}
\end{figure}

The following analysis is performed to determine whether the dark and light pixels in the MNIST experiment are discriminative. We plot histograms of all pixel values for each class separately. The histograms look very similar, as shown in Figure~\ref{fig:mnist_hist}, indicating that the dark or light pixels are not predictive alone. Therefore we add red or blue squares/variables which are both low variance and predictive. 

\label{var_mnist}

\subsubsection{Distribution of the weights in CT.}  
\begin{figure}[t!]
\centering
  \includegraphics[width=\textwidth]{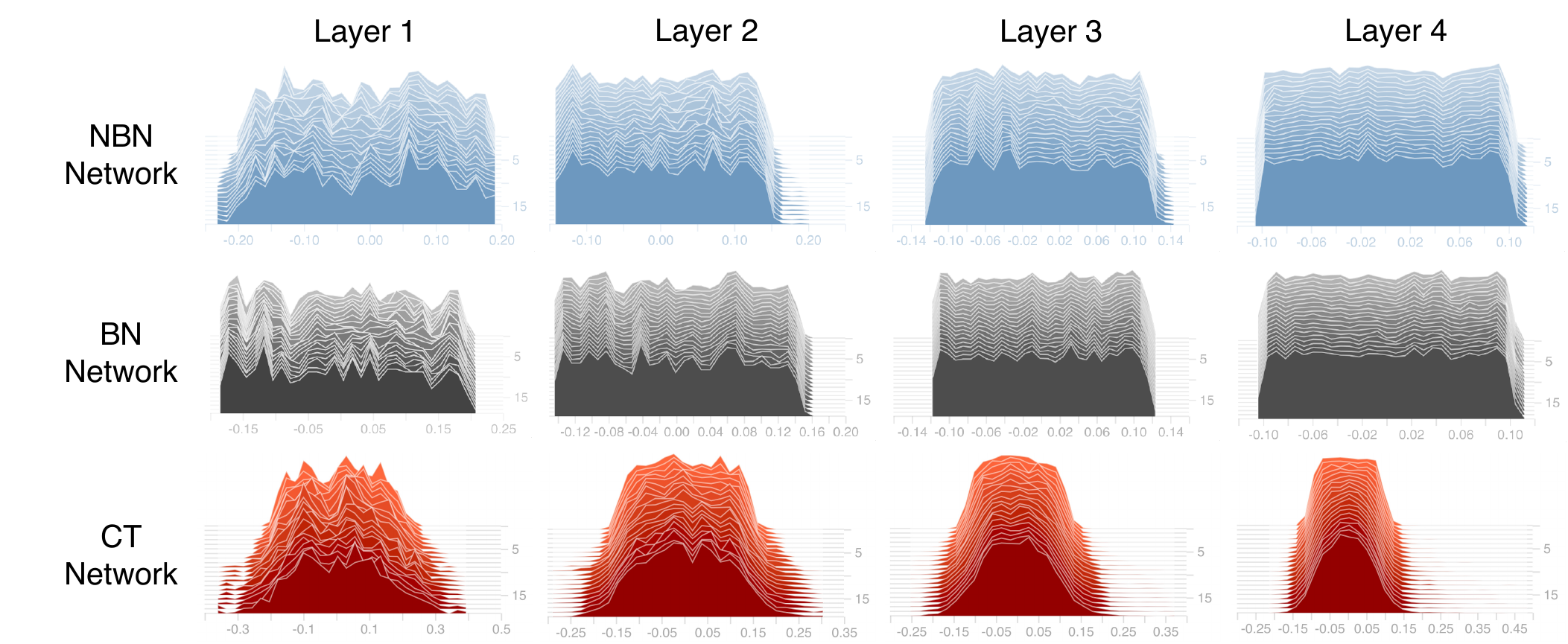}
  \caption{Visualization of 4 weight distribution for our MNIST classifier. Horizontal direction: magnitude of parameters. Orthogonal to page: epoch number. Up direction: histogram of parameters.
}
\label{fig:weight_hist}
\end{figure}

We provide visualizations of the distribution of NN weights both before and after removing the BN module on MNIST in Figure~\ref{fig:weight_hist}. Our CT network (especially on layer 1) progresses to adapt its distribution to the teacher’s distribution by increasing the histogram width to include weights with higher magnitudes (see the changes along the axis orthogonal to page). The same phenomenon holds for other layers, but the variation in weights decreases as the receptive field grows larger.

\subsubsection{ImageNet-C}
This dataset provides 15 algorithmically generated corruption types that are applied on the ImageNet validation set [Hendrycks et al. 2019]. These corruptions are drawn from four categories depicted in Tab. \ref{imagenetc_table}. In this task, the network is only trained on the original images, and then tested on the corrupted test sets. We observe a significant (3.6\%) improvement of mCE (mean Corruption Error), compared to the baseline (which uses BN layers in the architecture).

\begin{table}[t!]
\centering
\setlength{\tabcolsep}{3pt}
\resizebox{\textwidth}{!}{
\begin{tabular}{lcccccccccccccccc}
\hline
\textit{}                          & \multicolumn{3}{c}{\textit{\textbf{Noise}}} & \multicolumn{4}{c}{\textit{\textbf{Blur}}}         & \multicolumn{4}{c}{\textit{\textbf{Weather}}}       & \multicolumn{4}{c}{\textit{\textbf{Digital}}}      &              \\ \hline
\multicolumn{1}{l|}{Network} & Gauss.  & Shot & \multicolumn{1}{c|}{Imp.}  & Def. & Gls.  & Motion & \multicolumn{1}{c|}{Zoom}  & Snow  & Frost & Fog   & \multicolumn{1}{c|}{Bright} & Cont. & Elas. & Pix.  & \multicolumn{1}{c|}{JPG}   & \textbf{mCE} \\ \hline
\multicolumn{1}{l|}{Res50} & 79      & 80   & \multicolumn{1}{c|}{82}    & 82   & 90    & 84     & \multicolumn{1}{c|}{80}    & 86    & 71    & 75    & \multicolumn{1}{c|}{65}     & 79    & 91    & 77    & \multicolumn{1}{c|}{80}    & 80.1        \\
\multicolumn{1}{l|}{Res50+CT}      & 69   & 69 & \multicolumn{1}{c|}{70} & 80 & 91 & 82  & \multicolumn{1}{c|}{86} & 81 & 81 & 68 & \multicolumn{1}{c|}{62}  & 62    & 87 & 82 & \multicolumn{1}{c|}{80} & \textbf{76.5}         \\ \hline
\end{tabular}
}
\caption{We show our preliminary results (without tuning) below, which demonstrate a significant decrease in the error of our method over the baseline (ResNet-50) on ImageNet-C. \textit{mCE}: model error averaged on all corruption types (ImageNet-C validation).}
\label{imagenetc_table}
\end{table}

\subsubsection{Camelyon-17, WILDS}.
For real-world OOD experiments that do not require manual corruption of inputs (e.g., adding a red square), we include results on the Camelyon-17 dataset from WILDS~\cite{koh2021wilds}. The test accuracy for an ERM DenseNet-121 baseline w/ BN is $70.3 \pm 6.4$, while that of a DenseNet-121 w/o BN is $72.0 \pm 0.9$. CT gets the best test accuracy of $73.9 \pm 5.3$. 

\subsubsection{The Background Challenge}
We ran our CT model on the Background Challenge dataset (Xiao et al. 2020). The goal here is to understand how much a network relies on nuisance background features to perform the classification task. This dataset is a subset of ImageNet, comprised of multiple test sets such as (1) \texttt{mixed\_rand}, where the foreground from a particular image is overlaid onto a background of a different image; and (2) \texttt{mixed\_same}: where the background of a random image of the same class is overlaid onto the foreground. When trained with CT, the accuracy is improved on multiple test sets compared to the baseline. CT also reduces the overall gap in performance (\texttt{background gap} = \texttt{mixed\_same} - \texttt{mixed\_rand}) from 14.3\% to 8.1\%. This demonstrates that CT is less sensitive to nuisance (brittle) features that are specific to the background of an image. 

\begin{table}[h!]
\centering
\begin{tabular}{lccc}
          & \texttt{original} & \texttt{mixed-rand} & \texttt{mixed-same} \\
          \midrule
Baseline  & 96.3     & 75.6       & 89.9       \\
CT (ours) & \textbf{96.9}     & \textbf{81.8}       & 89.9      
\end{tabular}
\caption{Test accuracy on ImageNet-9 (IN-9L) for ResNet-50.}

\label{bg_challenge}
\end{table}

\subsection{Uncertainty evaluation metrics}
\label{uncertainty-app}
Calibration is defined as the degree of matching between the uncertainty predicted by the model and the true uncertainty in the data \cite{guo2017calibration}, and the Avg. Cal. Err. is defined as the average of the absolute difference between each sample's prediction confidence (the probability assigned to this sample by the model) and the model accuracy. 
For example suppose that we have 100 samples and our model assigns a label to each of the samples with the probability of 80\%. Now if the model predicts exactly 80 out of 100 samples correctly, we would say the model is completely calibrated and its Avg. Cal. Err. is 0. 
For Avg. Cal. Err. we report root mean squared error, mean absolute error, and miscalibration area. Miscalibration area is the area between the calibration curve and ideal diagonal \cite{tran2020methods} and the less this area is, the more calibrated the model is. We also report the mean absolute error and root mean square error for Adv. Cal. Err. for different group sizes, where group size refer to the proportion of test dataset. To calculate this, for each group size, we draw 10 random groups from the test dataset and calculate the mean absolute error and root mean squared error for each of them and report the worst one \cite{chung2021beyond}. Sharpness \cite{gneiting2007probabilistic} shows the concentration of the predictive distribution (i.e. probability distribution over classes). The Continuous Ranked Probability Score (CRPS) \cite{deisenroth2020mathematics} is the quadratic difference between the cumulative distribution function (CDF) of the predicted distribution and the empirical CDF of the observation. We observe that the baseline's predictions have lower accuracy compared to our method. Additionally, the baseline method suffers from a significant overconfidence in its predictions. CT, on the other hand, predicts with higher accuracy and less overconfidence.
\end{document}